\pdfoutput=1

\documentclass[11pt]{article}

\usepackage[final]{acl}

\usepackage{times}
\usepackage{latexsym}

\usepackage[T1]{fontenc}

\usepackage[utf8]{inputenc}

\usepackage{microtype}

\usepackage{inconsolata}

\usepackage{graphicx}

\usepackage{url}
\usepackage{booktabs}
\usepackage{makecell}
\title{RusCode: Russian Cultural Code Benchmark for\\ Text-to-Image Generation}

\author{
  \textbf{Viacheslav Vasilev\textsuperscript{1,2}},
  \textbf{Julia Agafonova\textsuperscript{1,3}},
  \textbf{Nikolai Gerasimenko\textsuperscript{1}},
  \textbf{Alexander Kapitanov\textsuperscript{4}},
\\
  \textbf{Polina Mikhailova\textsuperscript{4}},
  \textbf{Evelina Mironova\textsuperscript{1}},
  \textbf{Denis Dimitrov\textsuperscript{1,5}}
\\
\\
  \textsuperscript{1}Sber AI,
  \textsuperscript{2}MIPT,
  \textsuperscript{3}ITMO University,
  \textsuperscript{4}SberDevices,
  \textsuperscript{5}AIRI
\\
  \small{
    \textbf{Correspondence:} \href{mailto:vasilev.va@phystech.edu}{vasilev.va@phystech.edu}
  }
}

\begin{document}
\maketitle
\begin{abstract}
Text-to-image generation models have gained popularity among users around the world. However, many of these models exhibit a strong bias toward English-speaking cultures, ignoring or misrepresenting the unique characteristics of other language groups, countries, and nationalities. The lack of cultural awareness can reduce the generation quality and lead to undesirable consequences such as unintentional insult, and the spread of prejudice. In contrast to the field of natural language processing, cultural awareness in computer vision has not been explored as extensively. In this paper, we strive to reduce this gap. We propose a RusCode benchmark for evaluating the quality of text-to-image generation containing elements of the Russian cultural code. To do this, we form a list of 19 categories that best represent the features of Russian visual culture. Our final dataset consists of 1250 text prompts in Russian and their translations into English. The prompts cover a wide range of topics, including complex concepts from art, popular culture, folk traditions, famous people's names, natural objects, scientific achievements, etc.
We present the results of a human evaluation of the side-by-side comparison of Russian visual concepts representations using popular generative models.
\end{abstract}

\section{Introduction}

In recent years, text-to-image (T2I) generation models have achieved a high level of photorealism and comprehension of complex textual prompts \cite{dalle3, esser2024sd3, kastryulin2024yaartartrenderingtechnology, arkhipkin2024kandinsky30technicalreport, vladimir-etal-2024-kandinsky}. This has significantly expanded the potential for using them in various applications,  such as advertising, design, education, and art. As these models work with both visual and textual concepts, their operation is closely related to various aspects of human culture. The increasing popularity of generative systems available to users worldwide means that models need to understand text prompts containing specific elements from various cultures. However, as a general rule, these models are trained using large open datasets or data collected from the Internet. Due to the widespread influence of English-speaking popular culture, there is a lack of cultural understanding of other geographical, national, and social groups among generation models \cite{10376690, 10.1145/3593013.3594016, 10.1145/3613904.3641951}. These restrictions will undoubtedly lead to incorrect generation results for specific cultural concepts, a loss of user interest, and a limited applicability of the model for real-world tasks. In the worst-case scenario, this could lead to undesirable social outcomes, such as unintentional insults \cite{ghosh2024generative}, inciting hostility, spreading misinformation, and perpetuating stereotypes and social biases \cite{10.1145/3600211.3604711, Cho2023DallEval, 10.5555/3666122.3668580}. This is one of the main reasons for many concerns regarding the use of generative Artificial Intelligence (AI) in general \cite{weidinger2023sociotechnicalsafetyevaluationgenerative, 10.1145/3600211.3604722}.

As human culture is linked to language, similar challenges have been considered previously in natural language processing (NLP) \cite{hershcovich-etal-2022-challenges, NEURIPS2023_ae9500c4, cao-etal-2024-bridging}. However, the cultural awareness issue in visual generation tasks remains largely unexplored. We understand cultural awareness in the generation model as knowledge of the cultural code. We mean the cultural code as a diverse set of concepts that members of a particular social group or nationality regularly encounter. These concepts are often an integral part of a person's cultural background and are widely accepted for communication within specific communities \cite{Corner1980}. At the same time, these concepts may be unfamiliar or even incomprehensible to other people, as they can contain complex, metaphorical, and fantastical elements.

In this paper, we present the RusCode benchmark dataset for evaluating the quality of image generation based on textual descriptions that include concepts from the Russian culture. We conduct cultural analysis with the participation of experts from various fields of the humanities, such as history, literature, sociology, psychology, and philology. Based on these diverse perspectives, we create a list of 19 categories that cover various aspects of Russian culture. We aim to develop a system of concepts that will be easily understood by most native Russian speakers. As a result, we construct a dataset consisting of 1250 complex textual descriptions in Russian and English, which reflect the contextual use of many concepts from traditional and modern Russian culture. When creating the prompts, we took into account the opinions of 13 people from various backgrounds, professions, and age groups. These descriptions include historical, artistic, folkloric, natural, technical and other elements. We also associate a real reference image of a particular entity with each prompt. These images can be used to evaluate the generation quality for the elements of Russian culture with the participation of people who are unfamiliar with Russian culture in detail. This allows one to use our dataset to evaluate multicultural image generation models. We use the collected prompts to generate images using popular models such as Stable Diffusion 3 \cite{esser2024sd3}, 
DALL-E 3 \cite{dalle3}, Kandinsky 3.1 \cite{arkhipkin2024kandinsky30technicalreport, vladimir-etal-2024-kandinsky}, and YandexART 2 \cite{kastryulin2024yaartartrenderingtechnology}. The results of a human evaluation of the side-by-side comparison of these models provide insight into the current state of multicultural understanding in the modern state of T2I generation.

Thus, the contribution of our work is as follows:
\begin{itemize}
    \item We analyze the concept of the Russian visual cultural code and create a list of categories that represent the basic cultural background of a native speaker within the context of Russian culture;
    \item We present a RusCode benchmark dataset of textual descriptions of Russian cultural concepts that can be used to assess the cultural awareness of text-to-image models\footnote{Dataset is available here: \url{https://github.com/ai-forever/RusCode}};
    \item We report on the human evaluation results of the side-by-side comparison of 4 popular text-to-image generation models using collected prompts.
\end{itemize}

\section{Related Works}

\subsection{Cultural Awareness of Generation Models}

We define that a model has a high level of \textit{cultural awareness} if it can generate semantically correct results for text prompts that contain specific concepts related to a particular culture. Multicultural awareness involves understanding of linguistic and semantic features \cite{wibowo2023copal}, as well as correctly semantic matching of concepts from different cultures \cite{10.1162/tacl_a_00634}. Earlier, a tendency towards Western culture in generative models has been noted \cite{bhatia2024localconceptsuniversalsevaluating, 10.1145/3600211.3604711, berg-etal-2022-prompt}. As language is a conduit of culture \cite{ventura2023navigating}, many NLP studies have focused on the issue of cultural awareness. This includes the task of adaptive translation \cite{peskov-etal-2021-adapting-entities}, offensive language detection \cite{zhou-etal-2023-cultural, 10100717}, dialog systems operation \cite{cao-etal-2024-bridging} and other tasks \cite{hershcovich-etal-2022-challenges}. The development of visual-language models (VLM) has led to a transfer of cultural awareness issues for the multimodal architectures \cite{nayak2024benchmarkingvisionlanguagemodels}. This problem was considered in the context of the visual question answering (VQA) task \cite{10.1145/3590773, romero2024cvqaculturallydiversemultilingualvisual}, image-text retrieval and grounding \cite{bhatia2024localconceptsuniversalsevaluating}. With the addition of a new modality, the problem of cultural awareness has become more acute. For example, there are different levels of understanding of concepts from regional cultures among modern VLMs \cite{nayak2024benchmarkingvisionlanguagemodels}. 
In text-to-image generation, quality metrics have long focused on the aesthetics and photorealism of generated results, while ignoring cultural awareness \cite{kannen2024aestheticsculturalcompetencetexttoimage} Several studies have identified significant gaps in the level of multicultural awareness for the most popular T2I models. As far as we know, our work represents the first comprehensive approach to the issue of cultural awareness in relation to Russian culture in the T2I task.

\begin{figure*}[t]
\includegraphics[width=\linewidth]{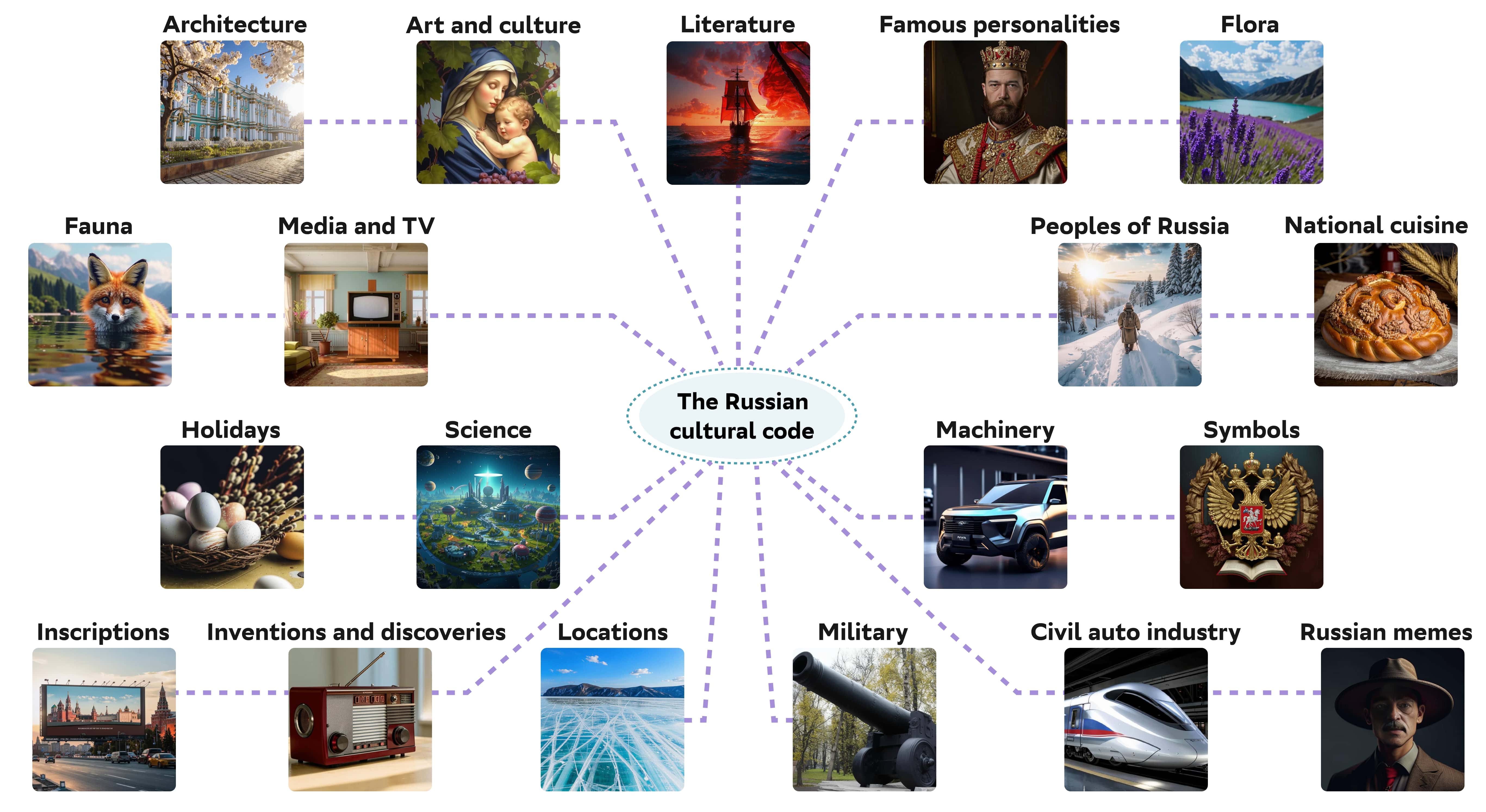}
  \caption{19 categories of Russian cultural code in our RusCode benchmark dataset. The images are generated by the Kandinsky 3.1 model \cite{arkhipkin2024kandinsky30technicalreport, vladimir-etal-2024-kandinsky}.}
  \label{fig:categories}
\end{figure*}

\subsection{Multicultural Benchmarks}

Benchmarks for evaluating the multicultural and multilingual abilities of generative models primarily emerged in NLP tasks. Due to the fact that most existing language models are designed primarily for English, several studies have focused on assessing the linguistic and grammatical features of other languages \cite{cahyawijaya-etal-2021-indonlg, https://doi.org/10.48550/arxiv.2208.13078, mukherjee-etal-2024-multilingual-text, kim-etal-2024-click, fenogenova-etal-2024-mera, taktasheva2024rublimprussianbenchmarklinguistic}. These benchmarks were designed to expand the range of multilingual knowledge tested in language models, as previously translated English-language datasets have overlooked various cultural and linguistic features \cite{kim-etal-2024-click}. This is especially important for common languages, which, nevertheless, have limited resources in terms of accessible open information on the Internet \cite{cahyawijaya-etal-2021-indonlg, https://doi.org/10.48550/arxiv.2208.13078}. The main types of tasks included in such benchmarks were question answering \cite{kim-etal-2024-click}, natural language generation \cite{cahyawijaya-etal-2021-indonlg}, multilingual dialog generation \cite{https://doi.org/10.48550/arxiv.2208.13078}, text style transfer \cite{mukherjee-etal-2024-multilingual-text}, and many other tasks \cite{fenogenova-etal-2024-mera}.

The next significant step forward was the development of multimodal benchmarks for evaluating multilingual VLMs \cite{NEURIPS2022_6a386d70, pmlr-v162-bugliarello22a, nayak2024benchmarkingvisionlanguagemodels, romero2024cvqaculturallydiversemultilingualvisual, inoue2024heronbench}. The range of benchmark tasks here primarily includes visual question answering (VQA) \cite{pmlr-v162-bugliarello22a, nayak2024benchmarkingvisionlanguagemodels, romero2024cvqaculturallydiversemultilingualvisual, inoue2024heronbench}, as well as cross-modal retrieval, grounded reasoning, and grounded entailment tasks \cite{pmlr-v162-bugliarello22a}. Among the findings regarding the results of applying these benchmarks, it has been noted that the quality of modern VLM models varies depending on geographic and cultural categories \cite{nayak2024benchmarkingvisionlanguagemodels}. In addition, efforts have been made to combine text collected from the Internet with text generated by a pre-trained image captioning model.

Due to the fact that the T2I task primarily requires an assessment of visual cultural characteristics, not much work has been done in this area. Currently, existing benchmarks are limited in terms of the number of languages and cultural categories they cover \cite{kannen2024aestheticsculturalcompetencetexttoimage}.In addition, they do not support the Russian language, despite its relatively high level of usage on the Internet\footnote{\url{https://w3techs.com/technologies/overview/content_language}}. In this work, we are, to the best of our knowledge, the first to conduct a comprehensive cultural analysis in order to create a benchmark dataset for assessing the quality of image generation incorporating elements of Russian culture.

\begin{table*}[!t]
\caption{The list of categories and subcategories in the RusCode benchmark dataset}   
\centering
\small
\begin{tabular}{ll}
    \hline
    \textbf{Categories} & \textbf{Subcategories} \\
    \hline
    Architecture & Orthodox Church; Sights; Major cities\\
    \hline
    Art and culture & Painting; Music; Theater; Ballet; Opera; Musical; Photography; Cinema; Cartoons;\\
    & Architecture; Sculpture; Decorative and Applied arts; Design; Circus\\
    \hline
    Literature & Folklore, fairy tales and legends; Poems; Prose; Fables; Children's literature\\
    \hline
    Famous personalities & Public figures; Cultural figures; Scientists; Entrepreneurs; Military; Cosmonauts;\\
    & Russian writers; Musicians; Actors; Bloggers; Politicians; Athletes\\
    \hline
    Flora & Coniferous plants; Deciduous plants; Trees and shrubs; Flowers and herbs; Tundra vegetation;\\
    & Steppe vegetation; Swamp vegetation; Desert vegetation; Fungi; Lower plants; Spore plants;\\
    & Fruit plants; Berries; Root crops\\
    \hline
    Fauna & Mammals; Fish; Birds; Reptiles; Cold-blooded; Artiodactyls; Ungulates; Carnivores;\\
    & Herbivores; Amphibians; Wild animals; Domesticated animals; Small animals; Large animals\\
    \hline
    Media and TV & Animation; Documentaries; TV Series; Talk Shows; Reality Shows; Feature films;\\
    & Social networks; Advertising\\
    \hline
    Peoples of Russia & Nationalities; Clothing; Traditions; Religion; Crafts\\
    \hline
    National cuisine & First courses; Second courses; Hot appetizers; Cold appetizers; Desserts; Meat dishes;\\
    & Fish dishes; Milk and dairy products; Bread and bakery products; Cereals; Vegetables;\\
    & Fruits; Soft drinks; Alcoholic drinks\\
    \hline
    Holidays & Religious holidays; Civil holidays; Political holidays; Family holidays; Professional holidays;\\
    & National holidays; International holidays\\
    \hline
    Science & Natural Sciences; Exact Sciences; Social and Humanitarian Sciences;\\
    & Fundamental Sciences; Applied Sciences\\
    \hline
    Machinery & Modern machinery; Soviet machinery; Agricultural machinery; Aviation machinery;\\
    & Shipping machinery; Construction machinery\\
    \hline
    Symbols & State symbols; National symbols\\
    \hline
    Inscriptions & Signage and billboards; Logos and symbols\\
    \hline
    Inventions and discoveries & \\
    \hline
    Russian memes & \\
    \hline
    Locations & Natural; Man-made\\
    \hline
    Civil auto industry & Passenger cars; Trucks; Public transport\\
    \hline
    Military technics & Tanks, armored personnel carriers, air defense; Airplanes and helicopters;\\
    & Ships and submarines; The rest; Equipment of the 19th century;\\
    & Equipment of the 17th-18th century; Equipment of an earlier period\\
    \hline
\end{tabular}
\label{tab:categories}
\end{table*}

\subsection{Ethics and Social Biases in Generative AI}

Insufficient cultural awareness of image generation models can lead to the spread of social biases, misinformation, and offensive content \cite{10.1145/3600211.3604711, Cho2023DallEval, 10.5555/3666122.3668580}. A number of studies have focused on reducing the biases in generative models that are based on factors such as race, skin color, gender, geography, and social status \cite{9656762, 10.1145/3600211.3604711, NEURIPS2023_ae9500c4, 10.1145/3630106.3658968, Clemmer_2024_WACV}. Cultural stereotypes were also seen as undesirable in favor of greater globalization \cite{berg-etal-2022-prompt, 10.1613/jair.1.15388}. Although we agree that cultural stereotypes can be offensive and need to be eliminated, knowledge about the specific cultural features should be retained by the model. For this reason, in our work we create a benchmark using expert knowledge to test the model's ability to capture real cultural features, while avoiding offensive stereotypes.

\begin{figure*}[t]
\centering
\includegraphics[width=\linewidth]{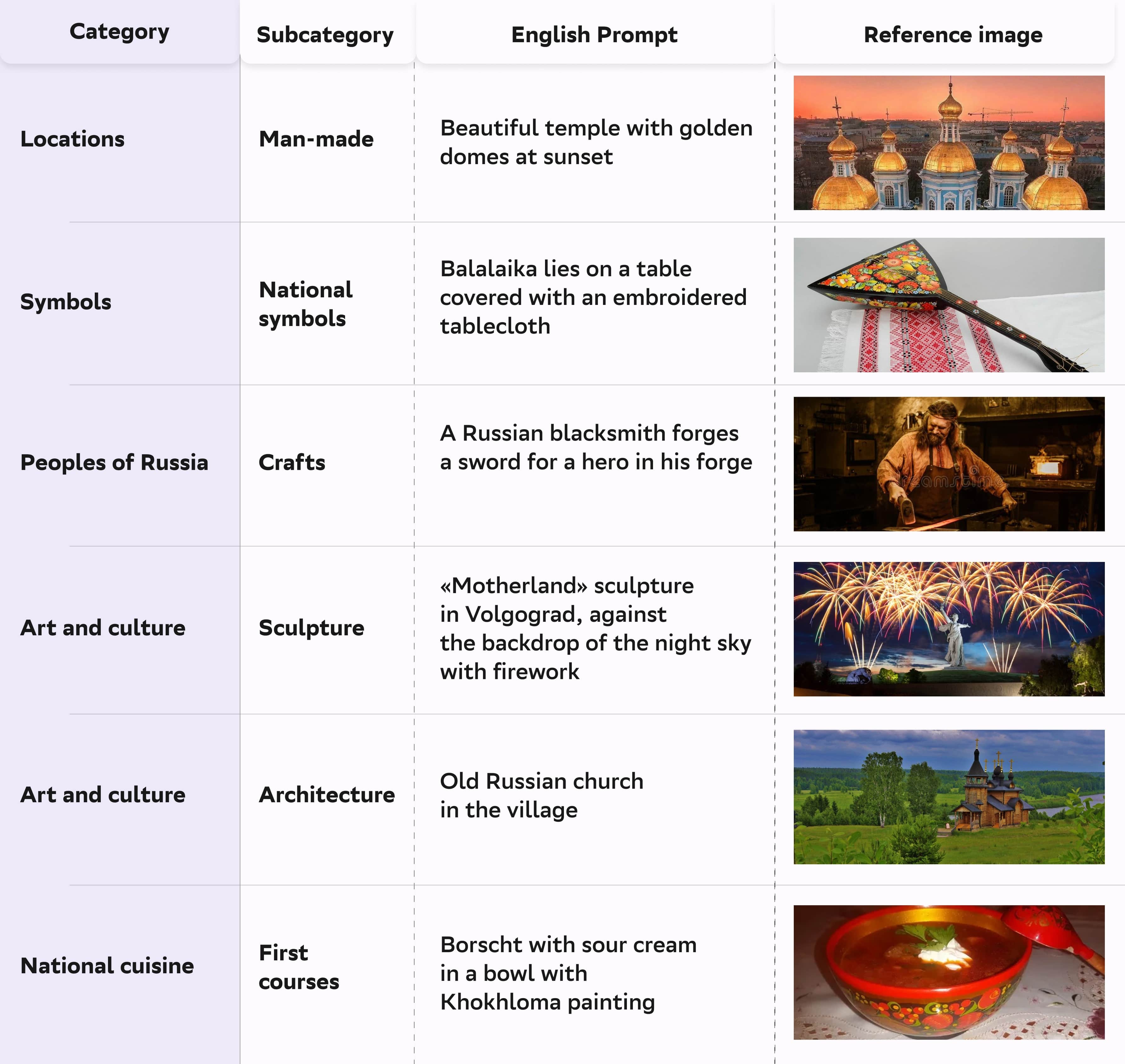}
\caption{Examples of prompts from RusCode dataset with corresponding reference images}
\label{fig:references}
\end{figure*}

\section{Cultural Code Analysis}\label{sec:analysis}

The concept of \textit{cultural code} is a complex idea that draws upon various fields such as history, cultural studies, sociology, philosophy, semiotics, and communication theory. The cultural code of a particular community is formed by symbolic systems such as language, art, as well as traditions, along with norms, values, social practices, and historical background. Popular culture, visual media and other forms of information significantly contribute to shaping the cultural code \cite{Corner1980}. Generation models need to be trained on a deeper understanding of cultural codes. This would improve the visual quality of their outputs, enhance their interpretative and communicative abilities, and enable the creation of systems that are sensitive to cultural diversity. This in turn would reduce biases and foster a more ethical approach in the content generation.

In this study, we explore the Russian cultural code. Drawing on existing research \cite{GOLOUBKOV2013107, billington2010icon, figes2002natasha, stites1992russian}, we highlight language, literature, art, religion, philosophy, folklore, and history as central components of Russian cultural identity. Since language reflects cultural characteristics \cite{https://doi.org/10.1525/eth.2002.30.4.401}, it is essential for the model to accurately interpret metaphors, proverbs, and figurative expressions that are common in everyday communication. From a visual standpoint, we also consider elements of contemporary popular culture, such as film and TV, as well as geographical places and natural objects. To ensure that our data selection aligns with this cultural framework, we consulted experts from different fields including history, literature, sociology, psychology, and philology. Their collective efforts have resulted identifying of 19 main categories and 125 subcategories, which are crucial for accurately representing the visual dimension of the Russian cultural code. Figure \ref{fig:categories} shows these categories with image examples. A complete list of categories and subcategories can be found in Table \ref{tab:categories}.

\begin{figure*}[t]
\centering
\includegraphics[width=0.9\linewidth]{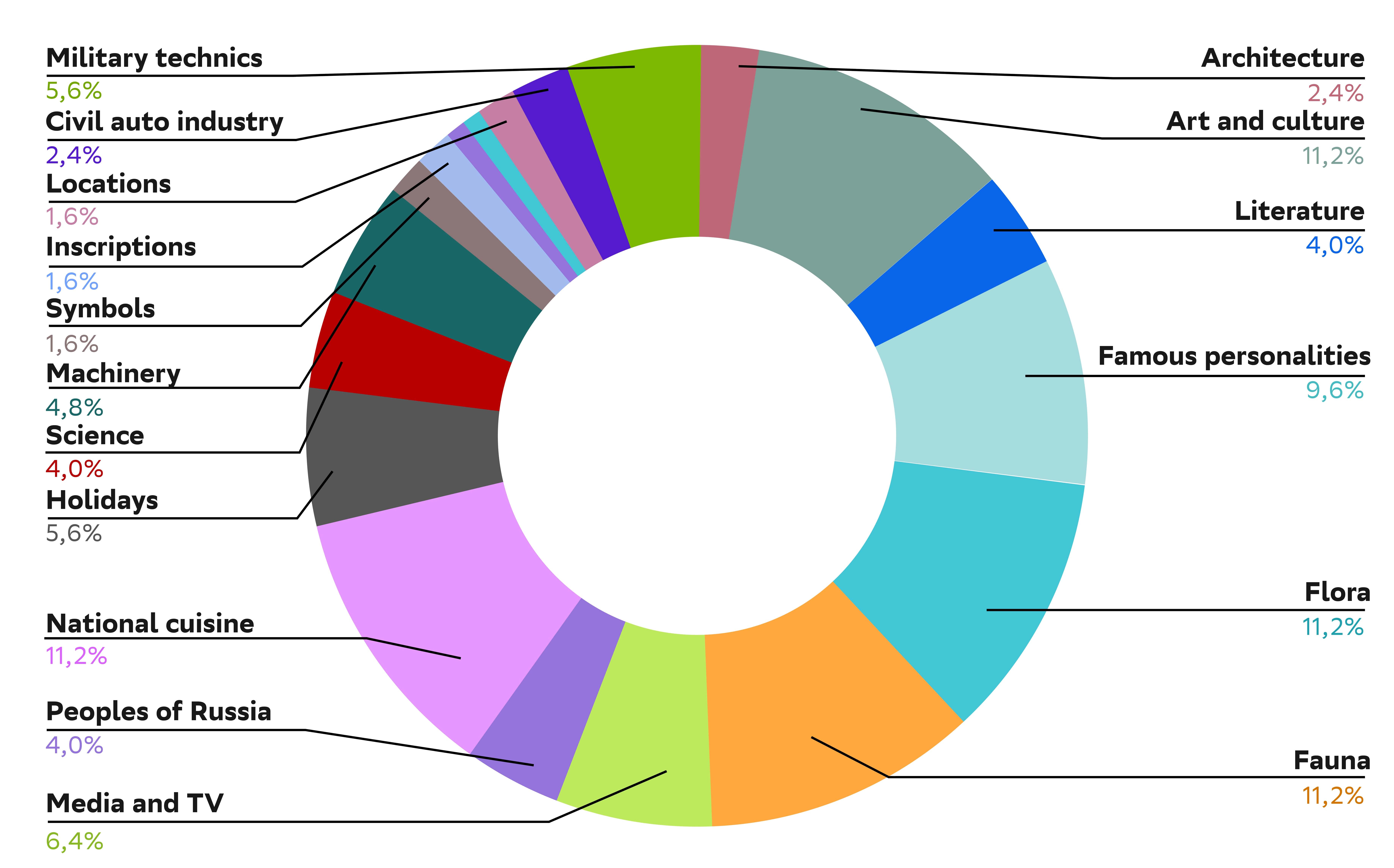}
\caption{The ratio of the number of collected prompts by each category in the RusCode dataset. }
\label{fig:diagram}
\end{figure*}

\section{Dataset}

\subsection{Prompts Creation}\label{sec:prompting}

\paragraph{General remarks.} After defining a list of main categories and subcategories that represent the Russian cultural code (Section \ref{sec:analysis}), we have created a dataset with prompts that correspond to these subcategories. We assigned ten complex prompts to each of the 125 subcategories in order to ensure a balanced distribution of cultural concepts in the dataset. Each prompt is presented in Russian and has an English translation variant. By a \textit{complex prompt}, we mean a textual description enclosing a certain cultural concept in a specific context of its use. For example, in the prompt \texttt{``art photography, aerial view of the Bolshoi Theatre in Moscow, evening, sunset''} two important entities for Russian culture related to each other are mentioned at once -- \texttt{``The Bolshoi Theater''} and \texttt{``Moscow''}. At the same time, the dataset contains concepts that are not directly expressed through visual images and require additional creative description. For instance, the discovery of an industrial process for producing synthetic rubber is represented in the dataset through the following description: \texttt{``Young Soviet chemist Sergei Lebedev made a discovery: a chemical reaction with a substance coming out of a flask, smoke and soot on the scientist's face, the invention of rubber''.} More examples of complex prompts are presented in Figure \ref{fig:references}.

\paragraph{Prompt-engineers.} It was essential for us that the prompts reflect the diverse experiences of people from the Russian culture. We have assembled a team of 13 prompt-engineers, including native speakers and professionals from various backgrounds. It includes a doctor, a manager, a cook, a pharmacist, a translator, an editor, a photojournalist, a psychologist, a car mechanic, a builder, a logistics expert, a linguist, and a copywriter. The age of people ranged from 19 to 46 years old. All prompt-engineers had previous experience in creating textual description datasets for generation models. They were officially employed when they completed their task and were aware of the work's objectives and the possible disclosure of their main areas of activity. Each team member has received instructions and been made aware of the rules for data collection, including ethical considerations and copyright laws.

\paragraph{Prompting.} We did not expressly limit the authors in any way at the initial stage of creating prompts. We recommended that they rely on their own experience and imagination. They were also provided with visual examples of how images with the Russian cultural code should look, descriptions of which they should create. Prompt-engineers actively used reference literature, books on painting and history, as well as open resources on the Internet. In order to avoid potential inaccuracies and enrich the dataset with more specific concepts, they did not utilize large language models. The final breakdown of the number of collected prompts by category is presented in Figure \ref{fig:diagram}.

\begin{figure*}[t]
  \includegraphics[width=\linewidth]{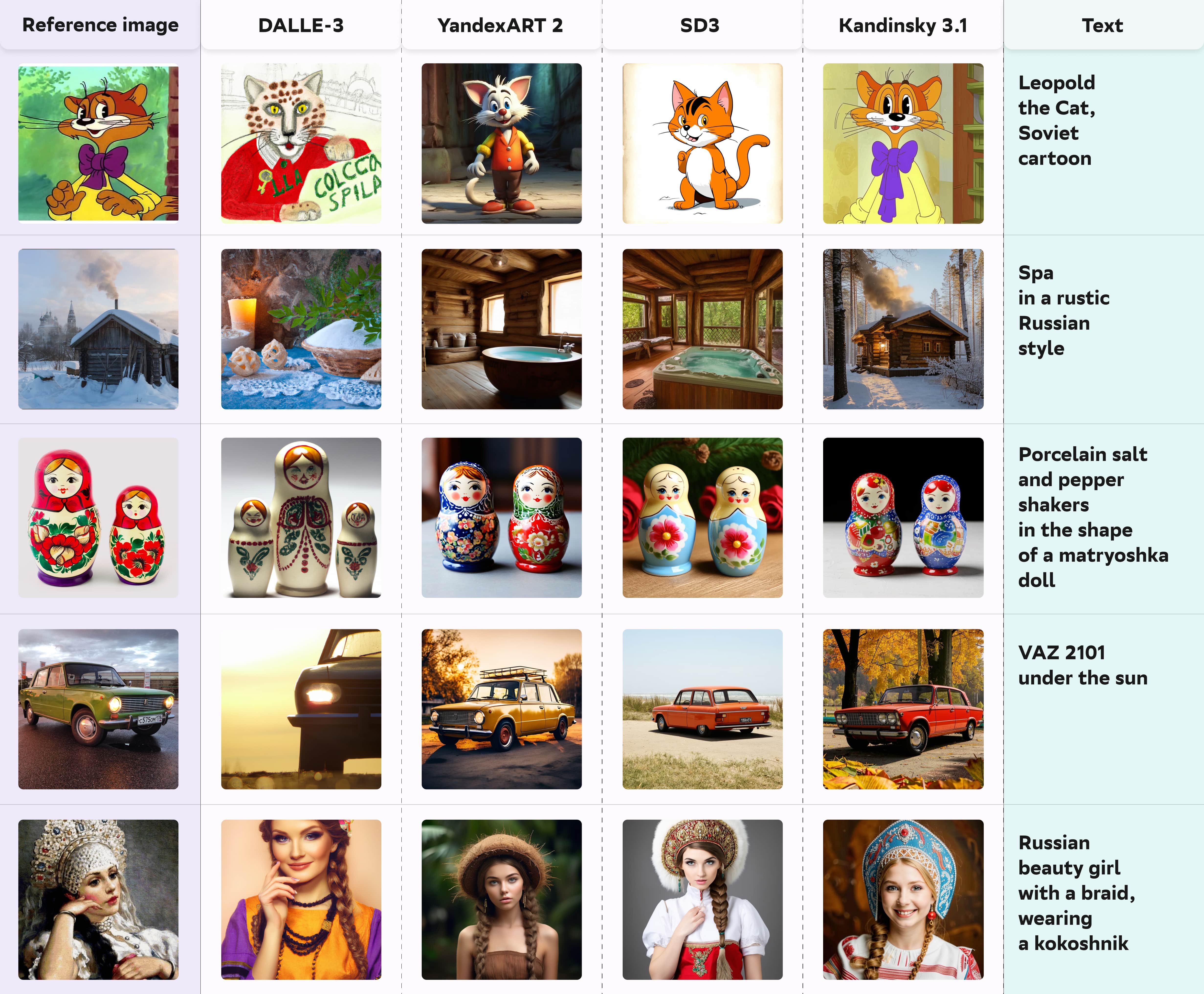}
  \caption{Comparison of Russian cultural code generations for popular text-to-image models. Reference is an example of a real image with a specific cultural concept from RusCode dataset.}
  \label{fig:examples}
\end{figure*}

\subsection{Prompts Filtering and Post-processing}

After the initial collection of prompts for each subcategory, they were selected and filtered by two experienced and professional prompt-engineers, who also contributed to the creation of a list of categories and subcategories (Section \ref{sec:analysis}). During the selection process, they were guided by their experience in creating high-quality and effective prompts, as well as by the idea of what prompts people use when they think about a particular entity and want to generate image with it. Additionally, professional prompt-engineers have corrected and rewritten the descriptions based on the results of popular queries in search engines related to Russian culture. They checked the correctness of the descriptions to ensure they matched reality and referenced literature, and added a plot to the prompts, enhancing its creativity, variety and detail. Special attention was given to prompts that describe complex and less popular topics, on which there is limited information in open sources or no visual content available. We also used the statistics on popular prompts for T2I generation models. Thus, 1250 prompts were selected from the 2500 initially created.

\begin{figure*}[t]
  \includegraphics[width=\linewidth]{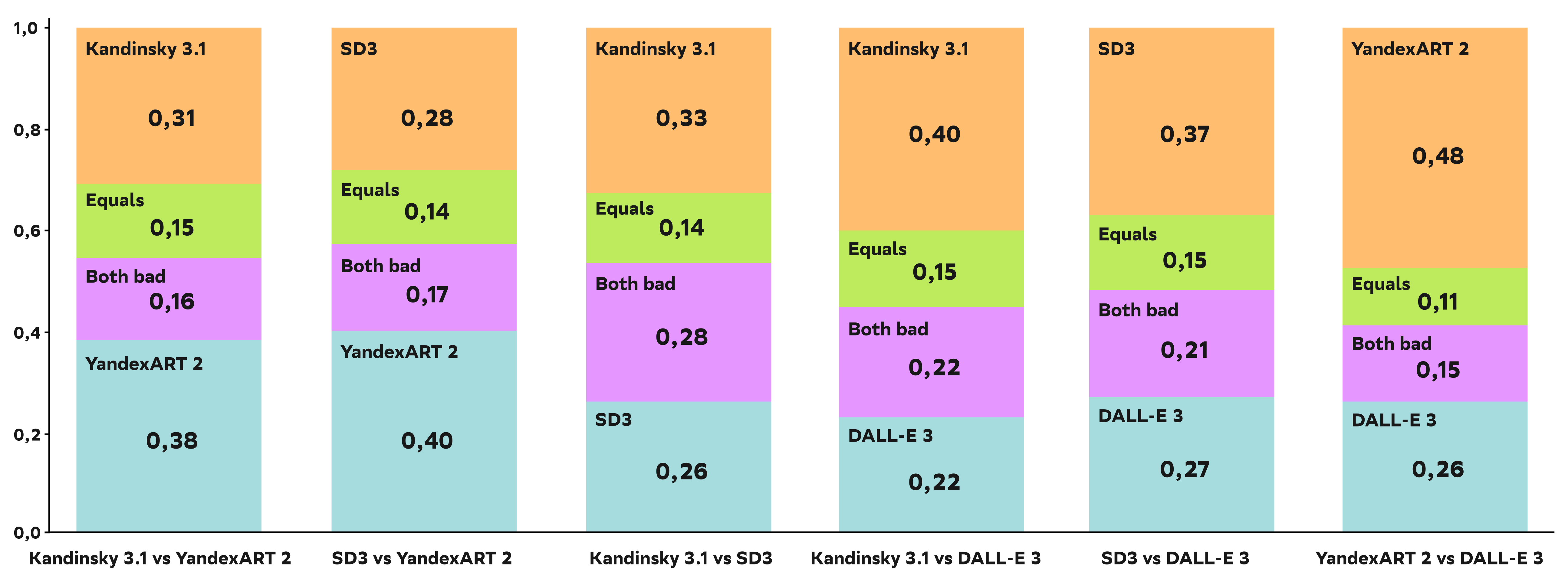}
  \caption{Human evaluation results of a side-by-side comparison between T2I model generations using text prompts from the RusCode dataset.}
  \label{fig:sbs}
\end{figure*}

\subsection{Reference Image Collection}

We also include high-quality reference images corresponding to each prompt in the dataset. These images are taken from open sources on the Internet. This is necessary to expand the possibilities of using our benchmark to visually assess the correctness of displayed cultural concepts with the help of people unfamiliar with Russian visual culture. By comparing the generated object or entity with the reference image, one can see how correctly the T2I model has performed in generating it. The images were selected after creating a set of prompts. Therefore, choosing images that matched the text descriptions as closely as possible was necessary. The reference images were selected by the same team of people who created the prompts (Section \ref{sec:prompting}). Special attention was paid to the aesthetics, photorealism, and overall visual quality of the images, including factors such as contrast, relative positioning of objects, brightness, color saturation, the naturalness of color reproduction, and sharpness. Examples of English-language prompts from different categories and subcategories, along with corresponding reference images, are presented in Figure \ref{fig:references}.

\section{Evaluation}\label{sec:evaluation}

\paragraph{Qualitative comparison.} We used the RusCode dataset to generate images using four popular T2I models, such as Stable Diffusion 3 \cite{esser2024sd3}, DALL-E 3 \cite{dalle3}, Kandinsky 3.1 \cite{arkhipkin2024kandinsky30technicalreport, vladimir-etal-2024-kandinsky}, and YandexART 2 \cite{kastryulin2024yaartartrenderingtechnology}. Figure \ref{fig:examples} shows several examples. As can be seen from the comparison with reference images, although all models have a fairly high level of visual quality, they cope with the generation of Russian cultural entities in different ways. Disadvantages can be expressed both in a lack of complete understanding of a particular entity, as well as in incorrect presentation of details. In some cases, models capture common features and produce examples of generalized concepts composed of individual recognizable elements, but they do not accurately reflect the essence of the reference.

\paragraph{Human evaluation.} We compared each of the four models side-by-side with the other three, conducting a human evaluation study. Each person was shown simultaneously the generations of two models without specifying their names. The task was to choose the image that most accurately matches the text description. A team of 48 people who were not involved in the creation of the dataset participated in the evaluation of the generated content. The age range of the participants was between 18 and 54 years. The fields of study and professions of the participants covered information systems, anthropology, programming, law, economics, philosophy, philology, linguistics, regional studies, political science, design, pedagogy, journalism, ecology, finance, sports, management, agriculture, and more. Each person viewed approximately 125 image pairs. The results of a general comparison across all categories are presented in Figure \ref{fig:sbs}. The results of comparing models in individual categories can be found in the Appendix \ref{sec:appendix_categories}. As can be seen, the Kandinsky 3.1 and YandexART 2 models significantly outperform the Stable Diffusion 3 and DALL-E 3 models. This indicates a lack of understanding of Russian culture among some of the most popular generative models. The Appendix \ref{sec:appendix_midjourney} contains the results of comparing the Kandinsky 3.1 and Midjourney v6 \cite{Midjourney} models.

It is important to note that sometimes models blocked generation for some prompts due to excessive self-censorship. For example, the YandexART 2 model blocked the prompts \texttt{``opera War and Peace, ball scene''}, \texttt{``Monument to the heroes of the Battle of Stalingrad on Mamayev Kurgan''}, \texttt{``Ostankino TV Tower at Night''}, \texttt{``Mikhail Gorbachev in a hat''}, etc. These and other prompts from our dataset do not contain any real offensive content. As a rule, automatic censorship does not react correctly to the mention of anything related to historical military topics or specific historical figures. When evaluating, we considered such censorship as a ``bad'' case. The Midjourney v6 model allowed us to use only 974 prompts out of 1250, so we did not include a comparison with it in the main text.

\paragraph{CLIP Score.} We used CLIP Score \cite{radford2021learningtransferablevisualmodels} to try to automatically assess the cultural awareness of the models on our dataset. The similarity score between the embeddings of the English prompts and the corresponding image embeddings are presented in Table \ref{tab:clip_score}. As can be seen, the results for all models are quite high and do not correlate with the human evaluation results. This confirms the inadequacy of using CLIP score for the cultural awareness assessment.

\begin{table}[!ht]
\caption{The similarity score between the embeddings of text prompts in English from the RusCode dataset and the embeddings of the corresponding generated images.}   
\centering
\small
\begin{tabular}{ll}
    \hline
     & CLIP Score $\uparrow$\\
    \hline
    Stable Diffusion 3 & 26.90\\
    \hline
    DALL-E 3 & 27.38\\
    \hline
    Midjourney v6 & 27.74\\
    \hline
    Kandinsky 3.1 & 26.89\\
    \hline
    YandexART 2 & 26.60\\
    \hline
\end{tabular}
\label{tab:clip_score}
\end{table}

\section{Discussion}

\paragraph{Taxonomy of errors.} We identified the features of errors that models encounter in trying to generate something from the Russian cultural code. In the absence of appropriate training examples, the model, even using a sufficiently detailed textual description, will not be able to correctly generate the necessary entity. Nevertheless, for large popular models, we observe a distortion of the entity or a display of an international concept rather than its replacement by an entity from another culture.

\paragraph{Automatic metrics.} As far as we know, there are currently no automatic metrics for assessing the cultural awareness of image generation models. The use of automatic metrics such as CLIP-score is not suitable for this task, since the evaluator model itself has a low level of cultural awareness (Table \ref{tab:clip_score}). This leads to the need to rely primarily on human evaluation, although we believe that our benchmark can lead to the development of automated tests for this task, for example, based on modern visual-language models, finetuned for cultural specifics.

\paragraph{Causes and mitigation of the cultural awareness gap.} We explain the advantage of the Kandinsky 3.1 and YandexArt 2 models by the presence of training data in the domain of Russian culture. The authors of Kandinsky 3.1 write about this explicitly in their technical report \cite{arkhipkin2024kandinsky30technicalreport}, while it is not exactly known for YandexArt 2. However, the fact that YandexArt 2 is primarily focused on interacting with Russian users allows us to make such an argumentative assumption. Following Kandinsky 3.1, we think that fine-tuning based on specific culture data will significantly improve the cultural awarness of the model. We also note that retrieval-augmented generation (RAG) methods \cite{NEURIPS2020_6b493230} can be productive in this direction.

\section{Conclusion}

In this paper, we proposed an open T2I benchmark dataset RusCode, which contains 1250 prompts in Russian and their corresponding translations into English. The dataset will be published under the MIT license. As far as we know, this is the first study in which the Russian cultural code has been examined in such a comprehensive and detailed manner. Despite the complexity of analyzing the national cultural code of any country, we have managed to create a system of categories and subcategories that accurately reflects the basic understanding of average users regarding prompts related to Russian visual concepts. The generation results of popular T2I models proof the existence of a cultural awareness issue, even though, in general, these models have some knowledge of generalized concepts. We strive to expand the use of our dataset. To do this, we attach reference images to the textual prompts, which can be used to mark the correctness of the generated entity. In the future, we aim to attract new experts and significantly increase our dataset, both in terms of the number of categories and the number of prompts in each subcategory. In the future, we also plan to expand this approach for video generation task \cite{arkhipkin2023fusionframesefficientarchitecturalaspects, 10815947}.

\section{Limitations}

\paragraph{Incompleteness of categories.} The assessment of the visual generation quality of elements of the Russian cultural code, which can be obtained using our dataset, may still not give a complete understanding of the capabilities of the generation model. This is directly related to the complexity and ambiguity of the concept of the Russian cultural code, which we significantly narrow down by providing a specific list of categories. This list could be significantly expanded, but we have focused on the most common concepts among ordinary users. At the same time, we do not exclude the fact that the dataset could contain data that requires more professional knowledge and goes beyond basic erudition or, for example, the school curriculum.

\paragraph{Insufficient representation of individual subcategories.} Within each subcategory, we have presented only 10 prompts to balance the data and avoid giving preference to any particular topic. At the same time, the importance of individual subcategories, such as ``Prose'' within the category of ``Literature'', is significant for Russian culture. It can be difficult to determine how the proportion of data in a dataset should reflect the social and cultural significance of a particular topic. Therefore, some relatively important concepts may be overlooked in the dataset, and preference may be given to less significant ones.

\paragraph{Ignoring more subtle differences.} In this paper, we use the term ``Russian cultural code'' to refer to a set of cultural phenomena that are common in Russia and the post-Soviet states. This may lead to some blending of differences between the various peoples and ethnic groups that inhabit the vast territory of Russia. However, our dataset is a significant step towards cultural awareness of these cultures, as the languages and cultural ideas of the people of Russia are closely related. As mentioned earlier \cite{cahyawijaya-etal-2021-indonlg}, training a model on closely related languages can enhance the quality of results in languages with limited resources. Therefore, we believe our dataset can also capture the cultural traits of the smaller ethnic groups.

\paragraph{Ambiguity and obsolescence.} In the humanitarian field, opinions can often lead to disagreements and disputes, and we acknowledge the possibility of mistakes related to cultural nuances. The cultural landscape is constantly evolving, influenced by elements of popular culture, new events, and trends, and we aim to monitor this process by regularly updating our data and expanding our team of experts. We will strive to standardize the process of prompts creation while maintaining the necessary level of freedom for prompt-engineers.

\paragraph{Quality assessment.} In this paper, we rely on human evaluation, based on a side-by-side comparison of model generation results for collected prompts. In the future, we plan to expand the list of quality metrics to include realism and visual quality assessment.

\paragraph{Recommendations for use.} Although we provide reference images, there may still be an incorrect match between the reference entity and the generated one when testing with people unfamiliar with Russian culture. We also recommend involving experts in relevant fields and using reference literature from trusted sources in such assessments.

\section{Ethical Statement}

\paragraph{Dataset Content.} We have avoided any potentially offensive or discriminatory concepts in our dataset. We do not include any racial prejudices or historical elements that might indicate a biased attitude towards any group in the concept of the cultural code. We strongly oppose nationalism and xenophobia. However, some elements of traditional culture might conflict with modern views of individuals or social groups. It is important to treat this with an understanding of their historical and cultural context.

\paragraph{Personal data.} Our dataset contains names and portraits of well-known historical figures of Russian culture. We would like to emphasize that we have not violated their privacy, as all information and images used in our dataset were obtained from open sources.

\paragraph{Usage.} Our research aims to promote multiculturalism and diversity in artificial intelligence. We oppose using our data for any illegal purposes, including incitement of hostility, hatred, or the creation of technologies that misinform, create false, or politically biased materials.

\paragraph{Payment for prompt-engineers and evaluators.} We properly paid the work of prompt-engineers who participated in the collection of the dataset (Section \ref{sec:prompting}), and the participants of the human evaluation study (Section \ref{sec:evaluation}). The average salary for each person exceeded the average salary in the city of his residence, according to publicly available government statistics.

\section*{Acknowledgments}

The authors express their gratitude to Tatiana Nikulina, Anastasia Alyaskina, Denis Kondratiev, Mikhail Kornilin, Julia Filippova, Irina Tabunova, Alexandra Averina, Stefaniya Kozlova, Andrei Filatov as well as Tagme and ABC Elementary Markup Commands.

\bibliography{main}

\appendix



\section{Side-by-side evaluation for Kandinsky 3.1 and Midjourney v6}\label{sec:appendix_midjourney}

\begin{figure}[t]
\centering
  \includegraphics[width=0.5\linewidth]{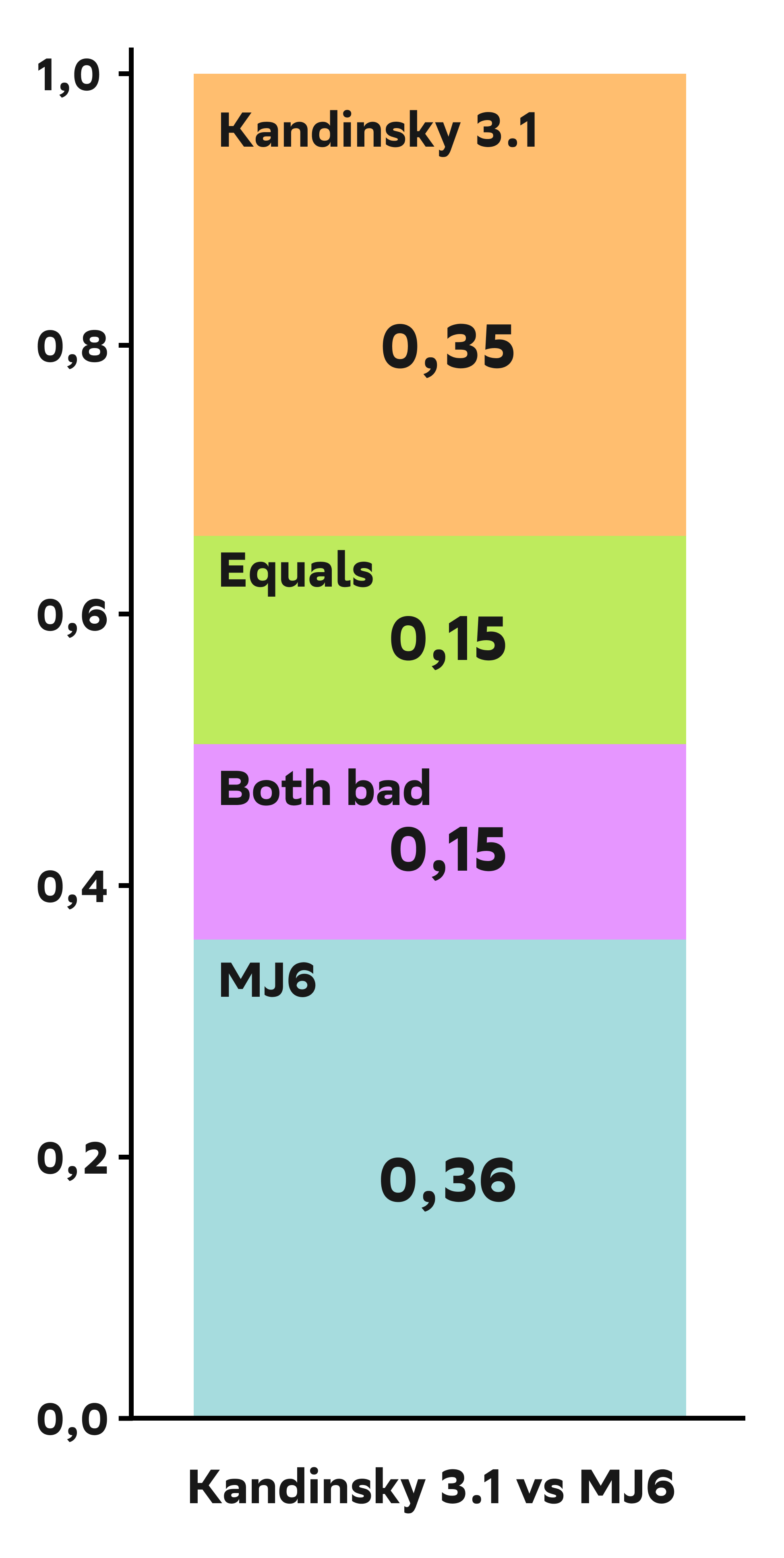}
  \caption{Side-by-side comparison between  Kandinsky 3.1 and Midjourney v6.}
  \label{fig:mj}
\end{figure}

According to the main results of the quality assessment, the YandexART 2 and Kandinsky 3.1 models were in the lead, but the YandexART 2 model often censored prompts and did not generate images. For this reason, we chose the Kandinsky 3.1 and additionally compared it with the Midjourney v6 model \cite{Midjourney}. As can be seen from the Figure \ref{fig:mj}, the models show competitive quality.

\section{Side-by-side evaluation by categories}
\label{sec:appendix_categories}

\begin{figure*}[htb]
\centering
  \includegraphics[width=\linewidth]{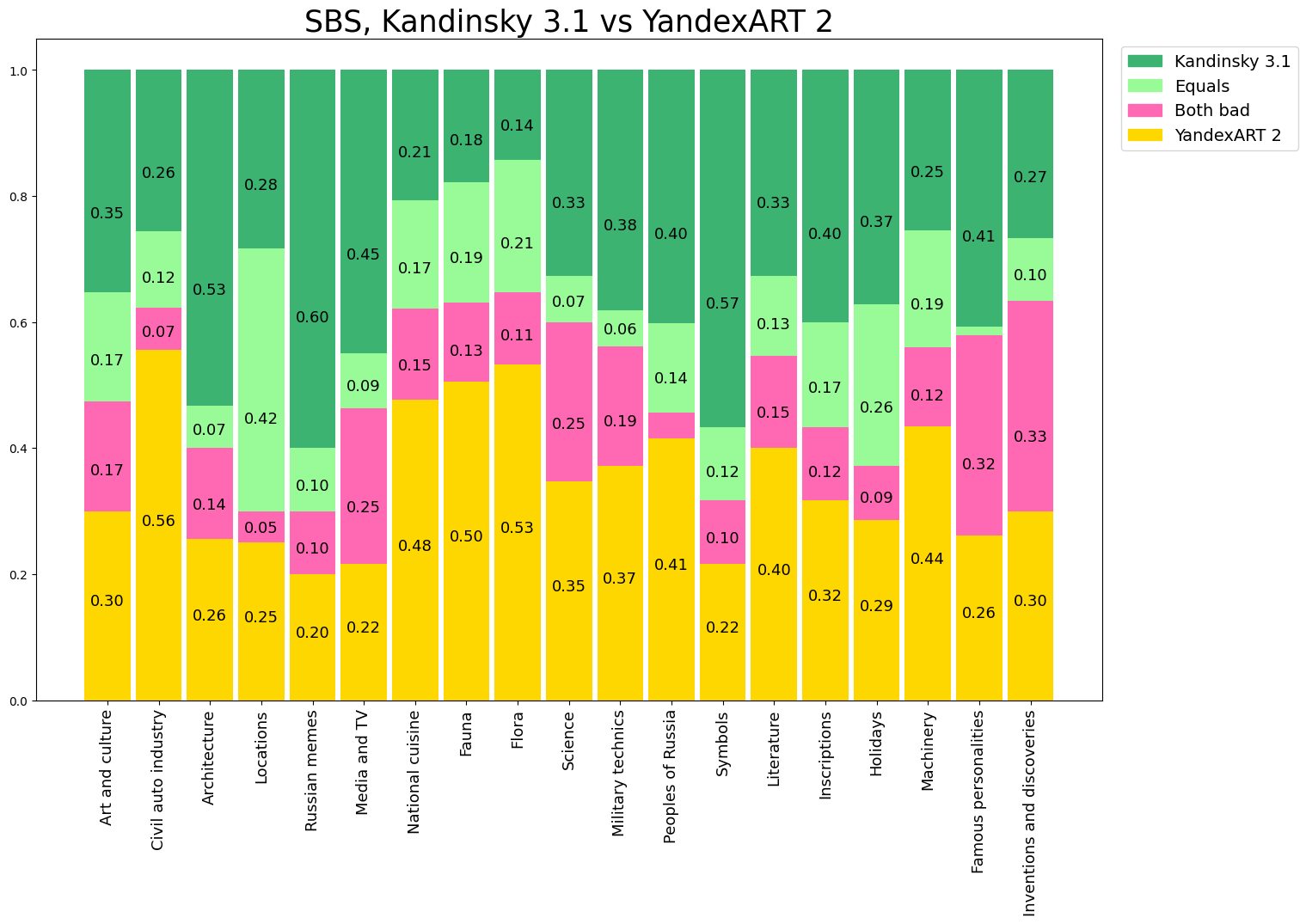}
\end{figure*}

\begin{figure*}[t]
\centering
  \includegraphics[width=\linewidth]{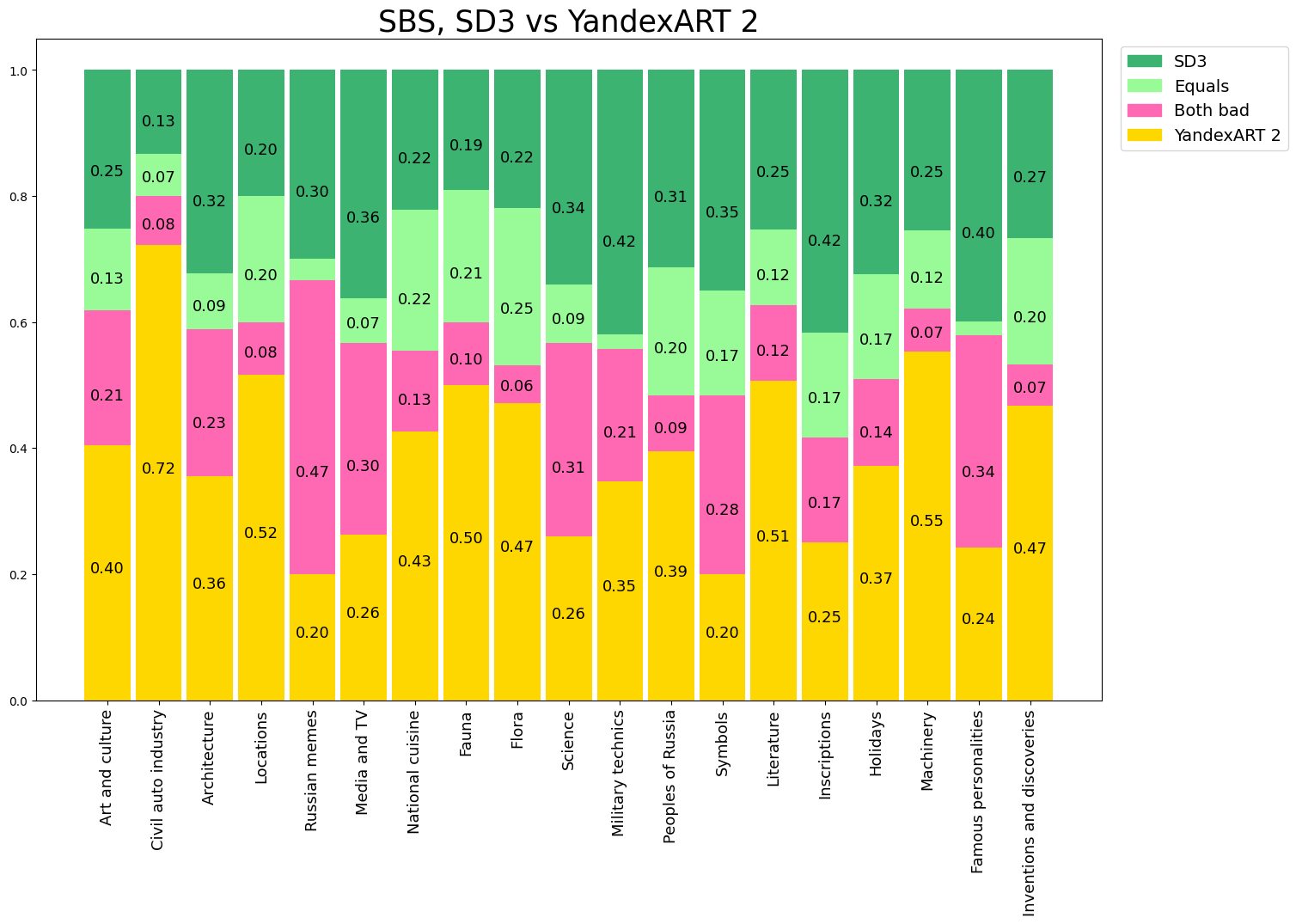}
\end{figure*}

\begin{figure*}[t]
\centering
  \includegraphics[width=\linewidth]{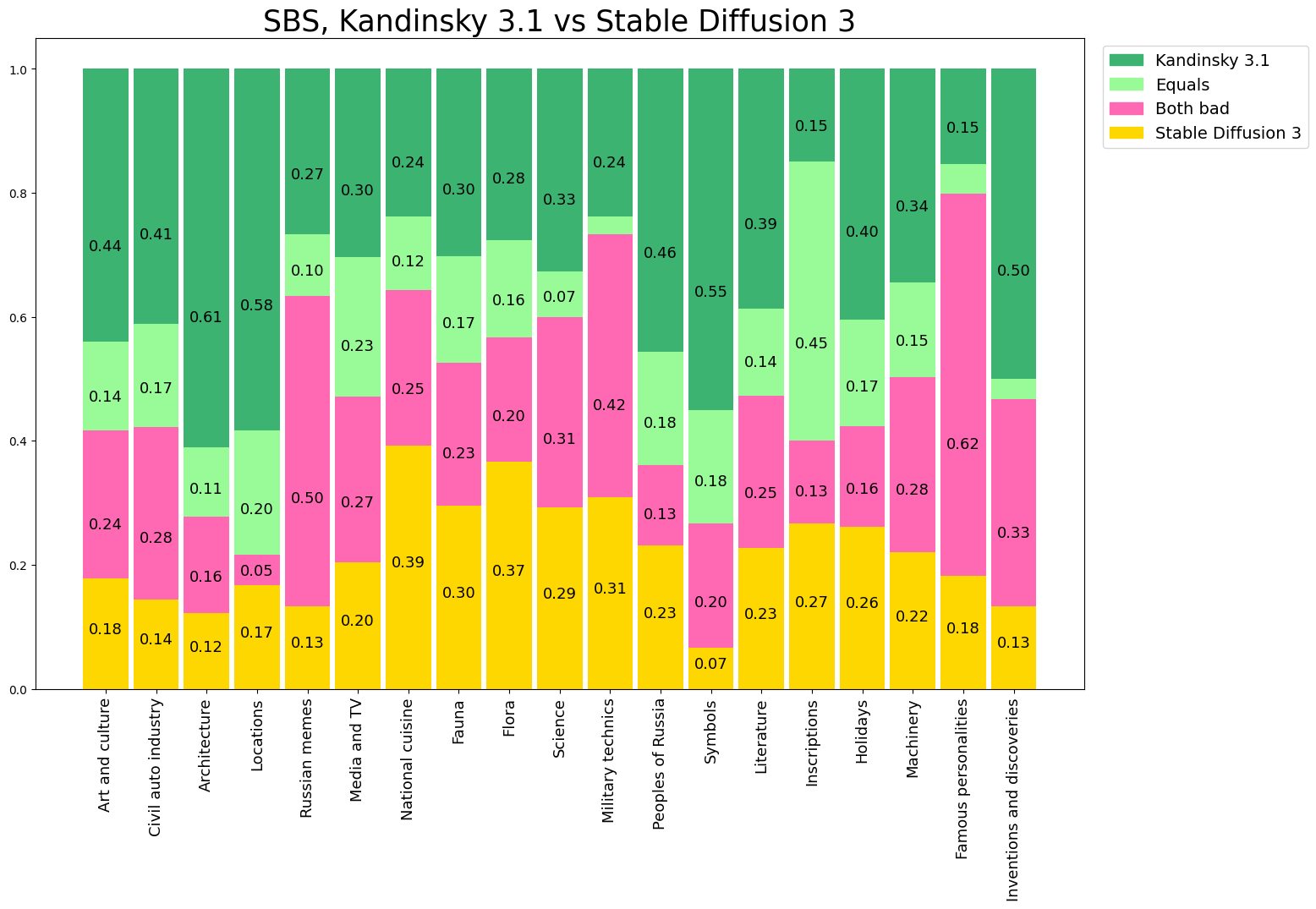}
\end{figure*}

\begin{figure*}[t]
\centering
  \includegraphics[width=\linewidth]{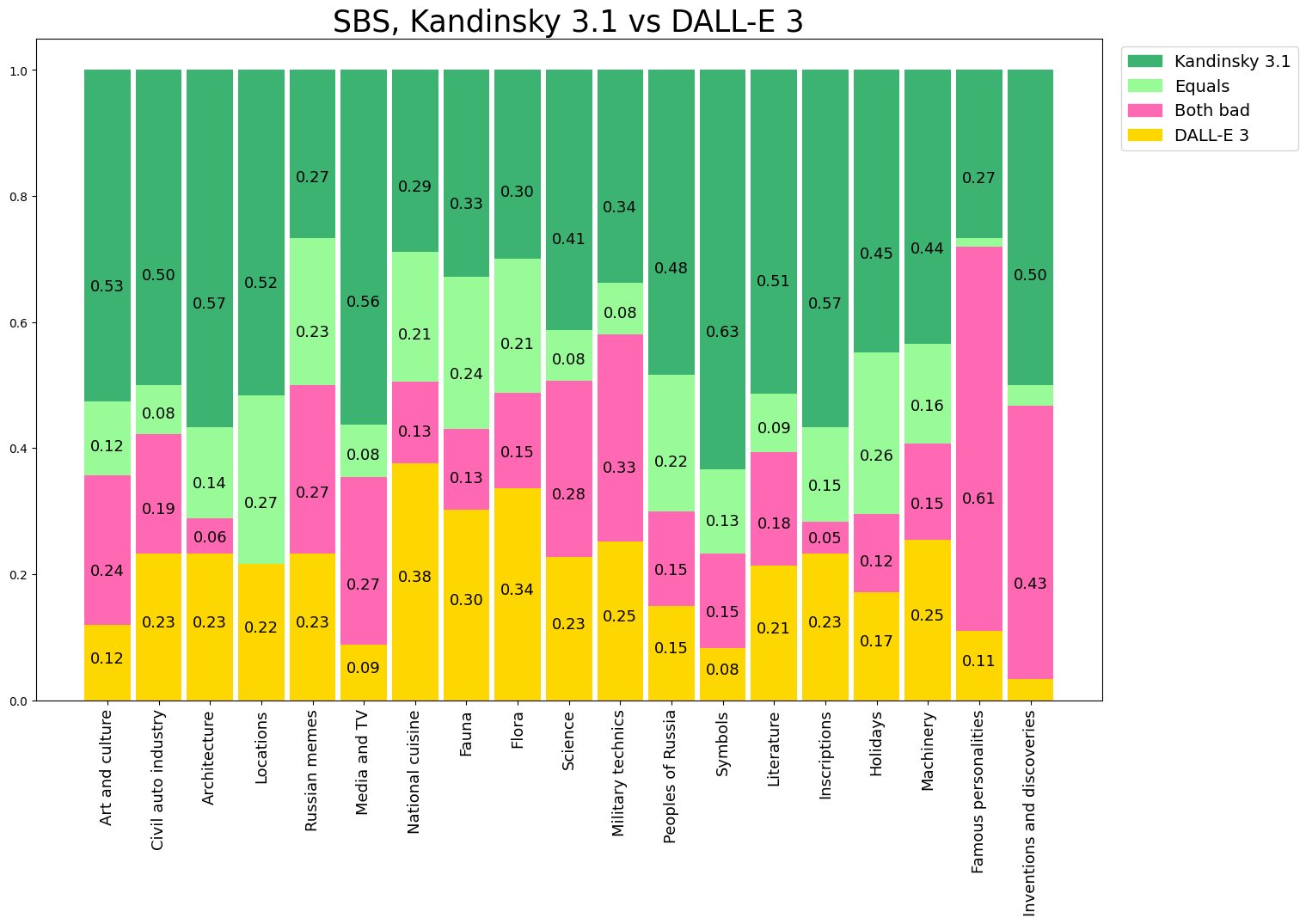}
\end{figure*}

\begin{figure*}[t]
\centering
  \includegraphics[width=\linewidth]{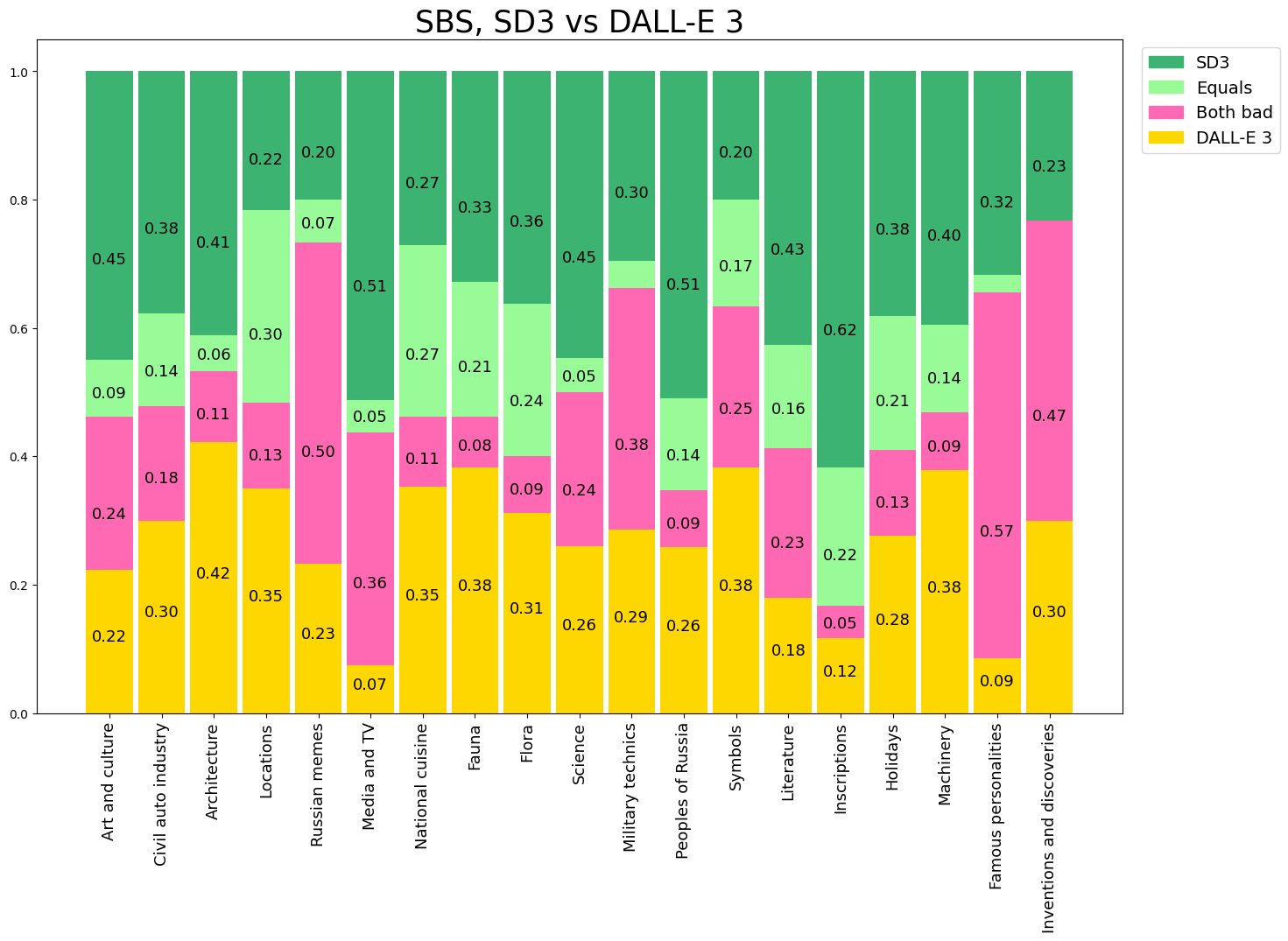}
\end{figure*}

\begin{figure*}[t]
\centering
  \includegraphics[width=\linewidth]{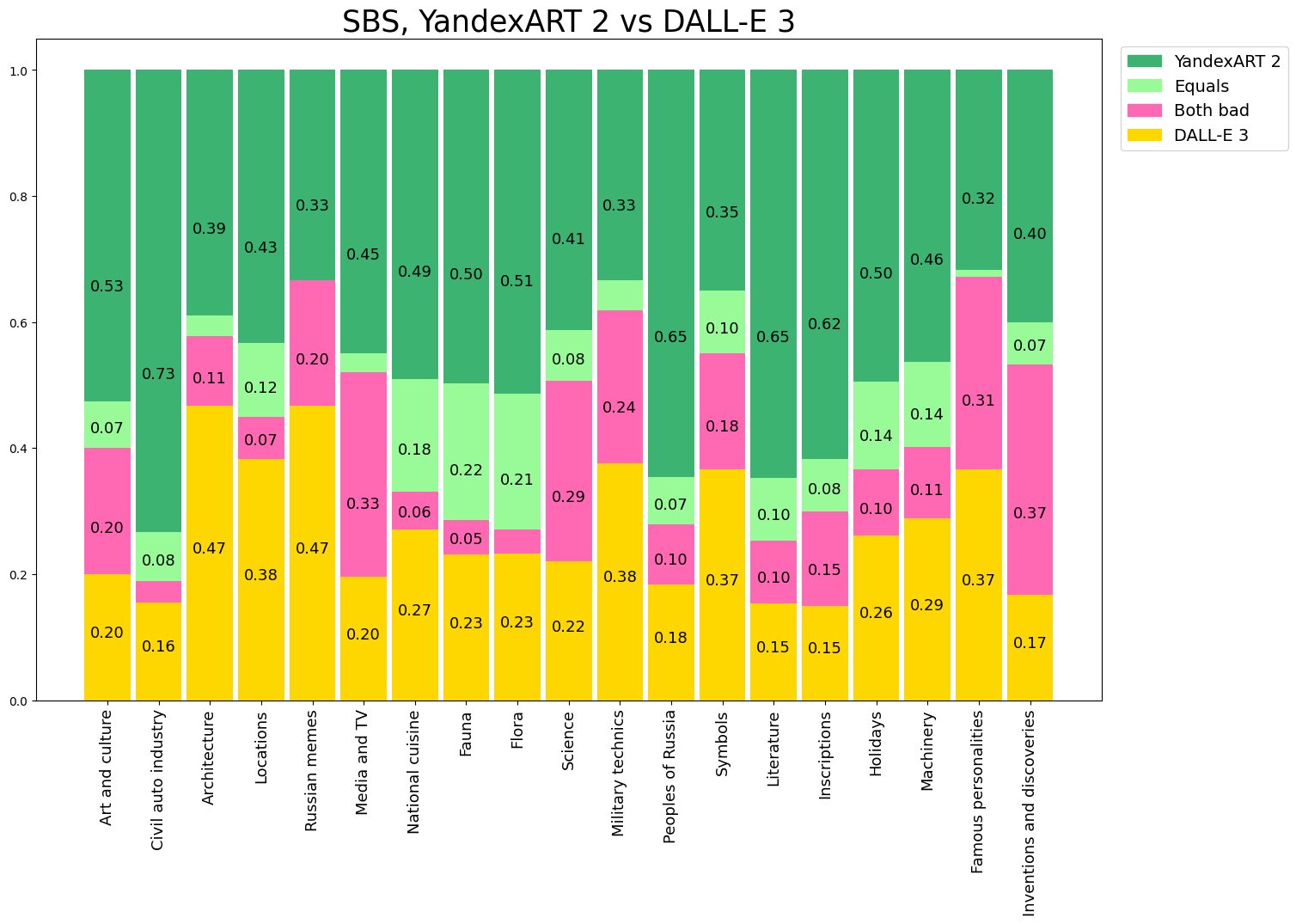}
\end{figure*}

\end{document}